\documentclass[letterpaper, 10 pt, conference]{ieeeconf}  

\IEEEoverridecommandlockouts                              

\usepackage{hyperref}
\usepackage{graphicx}
\usepackage{graphics} 
\usepackage{epsfig} 
\usepackage{times} 
\usepackage{authblk}
\usepackage{amsmath} 
\usepackage{amssymb}  
\usepackage{booktabs}
\usepackage{todonotes}
\usepackage{cleveref}
\usepackage{float}
\usepackage{siunitx}
\usepackage{makecell}

\title{\LARGE \bf
Autonomous Marker-Less Rapid Aerial Grasping
}

\author{
Erik Bauer$^{1}$, Barnabas Gavin Cangan$^{1}$, and
Robert K. Katzschmann$^{1}$
\thanks{$^{1}$Soft Robotics Lab, ETH Zurich, Switzerland}
\thanks{\tt \footnotesize 
\{\href{mailto:erbauer@ethz.ch}{erbauer},
\href{mailto:bcangan@ethz.ch}{bcangan},
\href{mailto:rkk@ethz.ch}{rkk}\}@ethz.ch}
}

\newcounter{daggerfootnote}
\newcommand*{\daggerfootnote}[1]{%
    \setcounter{daggerfootnote}{\value{footnote}}%
    \renewcommand*{\thefootnote}{\fnsymbol{footnote}}%
    \footnote[2]{#1}%
    \setcounter{footnote}{\value{daggerfootnote}}%
    \renewcommand*{\thefootnote}{\arabic{footnote}}%
    }

\begin{document}

\maketitle
\thispagestyle{empty}
\pagestyle{empty}

\begin{abstract}
In a future with autonomous robots, visual and spatial perception is of utmost importance for robotic systems. Particularly for aerial robotics, there are many applications where utilizing visual perception is necessary for any real-world scenarios. Robotic aerial grasping using drones promises fast pick-and-place solutions with a large increase in mobility over other robotic solutions. 
%
%
Utilizing Mask R-CNN scene segmentation (detectron2), we propose a vision-based system for autonomous rapid aerial grasping which does not rely on markers for object localization and does not require the appearance of the object to be previously known. Combining segmented images with spatial information from a depth camera, we generate a dense point cloud of the detected objects and perform geometry-based grasp planning to determine grasping points on the objects. 
%
In real-world experiments on a dynamically grasping aerial platform, we show that our system can replicate the performance of a motion capture system for object localization up to 94.5\% of the baseline grasping success rate. 
%
%
With our results, we show the first use of geometry-based grasping techniques with a flying platform and aim to increase the autonomy of existing aerial manipulation platforms, bringing them further towards real-world applications in warehouses and similar environments.\daggerfootnote{Code: \url{https://github.com/srl-ethz/detectron-realsense}}
\end{abstract}

\section{Introduction}

    \subsection{Motivation}
    Over the last five years, drones have gained significant popularity as tools to observe and capture the world around us through photographs and videos both for consumers and industry alike. With the rise of soft gripping technology, drones are gaining more capabilities for physical interaction with their environments. However, these capabilities will only find use if the drones have a way of efficiently perceiving the environment around them. The capabilities of an autonomous robotic system are always going to be limited by the information the system receives. For instance, a drone which can grasp objects but has no way of locating its target objects is rather limited for any real-world autonomous use cases. 
    
    A common solution in research is to use motion capture systems as a quick and precise solution to provide localization data. However, it is infeasible for industrial use to attach markers to every object that a robot should grasp. Solutions for object localization using fiducial markers like \textit{ArUco} markers~\cite{GarridoJurado2014AutomaticGA} exhibit the same problem. Vision-based target localization solutions that solely rely on monocular vision also do not have the capability of perceiving depth and cannot provide us with sufficient spatial information for rapid grasp planning. 
    
\begin{figure}[t]
              \centering
              \includegraphics[width=0.8\linewidth]{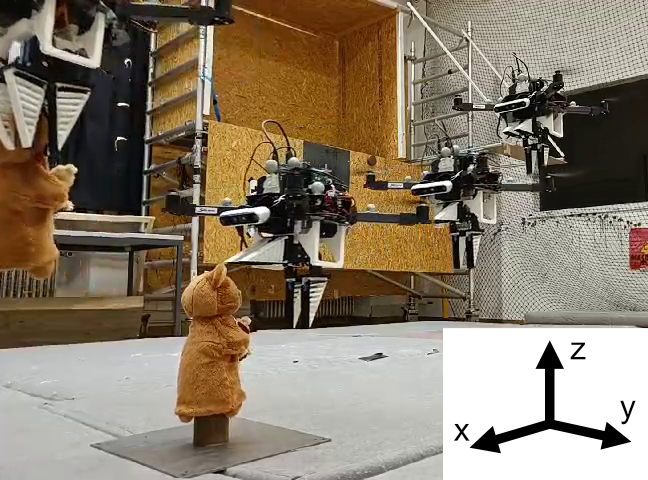}
              \\
              \centering
              \caption{Our proposed real-time scene segmentation and geometry-based grasp planning enables rapid aerial grasping (\SI{3}{\second} swoop duration) to pick up a target object using its soft gripper. We eliminate the need for artificial markers on grasp targets and perform grasp planning using an extracted point cloud of the object.}
              \label{drone_in_flight}
    \end{figure}

    To solve these challenges, we present a vision-based, marker-less system for object localization and grasp planning. Using a stereoscopic camera (\textit{Intel RealSense D455}~\cite{IntelRealSense}) for color and depth information, Mask R-CNN-based~\cite{8237584} scene segmentation (\textit{detectron2}~\cite{wu2019detectron2}), and efficient point cloud processing for grasp planning (\textit{Open3D}~\cite{Zhou2018}), our system gives robots the ability to perform marker-less autonomous grasping. We focus on aerial grasping by rapid swooping motions (\Cref{drone_in_flight}), which is accompanied by major challenges such as a substantial increase in tracking uncertainty, making it more challenging to execute precise grasps. Furthermore, the grasp execution starts from long distances, which reduces the pixel size of the objects in the image and therefore increases the importance of precise scene segmentation.
    
    We use \textit{RAPTOR}~\cite{Appius2022RAPTORRA} as a platform for deploying our system. \textit{RAPTOR} combines a quadcopter with a \textit{soft Fin Ray \textsuperscript{\textregistered} gripper} which passively adapts to the shape of the objects that are being grasped. It is a platform that is capable of dynamically picking up different objects at high speeds. Using \textit{RAPTOR} as a platform also allows us to compare our grasping performance against a known baseline grasping performance using a motion capture system for object localization.
    
    Our solution allows a mobile aerial platform to quickly and autonomously detect possible grasping targets (without any markers attached to them), localize them, and execute a grasp\footnote{Video attachment: \url{https://youtu.be/6hbhAT4l90w}}. Using Mask R-CNN~\cite{wu2019detectron2} image segmentation, we can generate dense point clouds of novel objects. 

    As opposed to previously used learning-based methods, our approach allows for highly precise object segmentation, even at larger distances (for example: bottles with 10 pixels width, about \SI{2.25}{\meter} away from the camera). This capability presents a key step in bringing mobile aerial robotic platforms, such as \textit{RAPTOR}, towards autonomous grasping and manipulation and eventually enable real-world applications.  

    Furthermore, we expand on the detection and grasp planning scheme that is typically shown in previous works with details such as a detailed runtime analysis and quantify the localization accuracy and grasping performance of our system. Finally, we also explicitly decouple our segmentation and grasp planning pipeline from the motion planning of the system. Thus, it will be easy to port the vision system to other mobile manipulation platforms for usage beyond the \textit{RAPTOR} platform. 
    
    \subsection{Related Work}
    \subsubsection{Aerial Grasping and Vision-based Control of Quadcopters}
    Control using visual information is generally referred to as visual servoing. Position-based visual servoing for quadcopters uses visual information to estimate a kinematic error in Cartesian space. Image-based visual servoing relies exclusively on estimating a control error in the two-dimensional image space. So-called hybrid approaches combine both visual servoing approaches into one control strategy. 
    
    Multiple works on aerial grasping exist that use marker-based target designations. Typically, they use reflective balls mounted on the object for localization using external infrared cameras (motion capture systems) or fiducial markers (\textit{ArUco} markers~\cite{GarridoJurado2014AutomaticGA}, \textit{AprilTags}~\cite{Olson2011AprilTagAR}). Using traditional rigid manipulators, Lippiello et al.~\cite{Lippiello2016HybridVS} and Buonocore et al.~\cite{Buonocore2015HybridVS} have shown a marker-based hybrid visual servoing approach for manipulation. More recently, rapid aerial grasping using soft gripping mechanisms has been shown using a motion capture system for estimating the pose of the target object by Fishman et al.~\cite{Fishman2021ControlAT, Fishman2021DynamicGW} and Appius et al.~\cite{Appius2022RAPTORRA}. 

    Marker-less methods for vision-based control of quadcopters have the advantage of not requiring target objects to be manually outfitted with markers. Luo et al.~\cite{Luo2020NaturalFV} perform feature matching for detection with a monocular camera while Lin et al.~\cite{Lin2019AutonomousVA} propose learning-based approaches for object detection and grasp planning. Ram\'on-Soria et al.~\cite{RamnSoria2020GraspPA} show a marker-less aerial grasping approach with a stereoscopic camera. They use a CNN (convolutional neural network) for detecting possible target regions that contain the known target object and extract the pose of the target object by applying the iterative closest points (ICP) algorithm~\cite{Besl1992AMF}. Ram\'on-Soria et al.~\cite{Soria2017DetectionLA} also show aerial grasping using traditional visual features like SIFT~\cite{LoweDavid2004DistinctiveIF} or FAST~\cite{Rosten2008FasterAB} for detecting objects.  A similar problem to grasping is perching: Thomas et al.~\cite{Thomas2016VisualSO} have shown a control approach for perching on known cylindrical objects using visual-servoing. Seo et al.~\cite{Seo2017AerialGO} show aerial grasping of similar cylindrical objects, making use of RANSAC (Random Sample Consensus) to detect the target cylinder and then use stochastic model predictive control for executing grasps.  
    

    
    In contrast to existing works, we use use a more powerful detection approach that is able to generalize to many different objects of unknown shapes. Using a learning-based method lifts requirements like having markers on the target objects or needing prior information about the shape or dimensions of the target objects as is the case for most feature-based methods. Mask R-CNN as segmentation method allows us to significantly expand the workspace of the robot compared to previous works, which rely on being within close range of the target object. Not only is this crucial for real-world applications, but it also allows us to use faster, swooping grasping maneuvers compared to the marker-less state of the art platforms. 
    
    \subsubsection{Vision-based Grasping With Statically Mounted Manipulators}
    Previous works show a number of approaches for grasp planning  with statically mounted manipulation platforms such as robotic arms that remain at a fixed position in their workspace. These approaches are mainly constructed around using vision to extract a point cloud of the object. One of the proven strategies is geometry-based grasp planning~\cite{ZapataImpata2019FastGC, Haschke2021GeometryBasedGP}. While in recent years learning-based approaches~\cite{Jiang2011EfficientGF, Mousavian20196DOFGV} for grasp planning have shown to be capable of generating precise grasps, they require comprehensive data sets. Methods using reinforcement learning~\cite{Levine2016EndtoEndTO} have also been explored, however, they have a high complexity in implementation and their generalization remains a challenging task. 
    
    While robotic arms and similar platforms have been used with great success for object manipulation, their inherently limited workspace restricts their practical applications. With our approach, which is based on extracting dense point clouds of objects, we lay the groundwork for implementing these successful grasp planning algorithms on highly mobile platforms such as \textit{RAPTOR}.  
    
    \subsection{Contributions}
   Our proposed vision system uses Mask R-CNN for powerful learning-based instance segmentation and extracting point clouds of target objects. We combine learning-based target point cloud extraction with geometry-based grasp planning, which enables marker-less rapid aerial grasping without the need for artificial markers on the target objects. 

   We contribute an extensive overview of the proposed system architecture and go over key insights for real-time aerial grasp planning. Included in our discussion are considerations regarding the choice of compute platform (onboard or offboard), instance segmentation framework, RGB-D video streaming infrastructure, process synchronization and inter-process communication. 
   
   Our method can precisely segment objects even at further distances which expands the workspace of the aerial platform and allows for rapid flights towards the target that do not rely on slow closed-loop visual servoing during the approach. 

    In real-world experiments using the \textit{RAPTOR}~\cite{Appius2022RAPTORRA} platform, we show that our system maintains grasping success compared to using a motion capture system for object localization and grasp planning. Furthermore, we highlight the vision-based object localization accuracy compared to object localization through a motion capture system and place it into context with previously used object pose estimation methods. 

    Finally, we outline future work to increase the abilities of vision-based aerial manipulation platforms. 
   

\section{Object Segmentation and Grasp Planning}
\subsection{Creating the Object Point Cloud}
There are two major steps required for acquiring point clouds of target objects: First, we use an RGB image of the object to detect it and use instance segmentation to get a segmentation mask. Second, we use the segmentation mask to filter out points in the RGB-D image and keep a dense point cloud of the target object.
\subsubsection{Instance Segmentation} We use Mask R-CNN (\textit{detectron2}~\cite{wu2019detectron2}) for segmenting the RGB images from the camera. R-CNN-based architectures~\cite{He2017MaskR, wu2019detectron2} have proven themselves to handle well small objects in the frame compared to more lightweight architectures~\cite{Bolya2019YOLACTRI, Wang2022YOLOv7TB} that also feature instance segmentation capabilities. Being able to detect and segment objects precisely from further distances is crucial to expand the workspace of the robot and rely less on human operators to guide it within close proximity of target objects. With a detection method like \textit{YOLOv5}~\cite{Jocher_YOLOv5_by_Ultralytics_2020}, we were able to reliably detect small objects like bottles at distances of around \SI{0.6}{\meter} whereas using Mask R-CNN allowed us to expand the distances up to \SI{2.25}{\meter}. 

While it is also possible to use simpler object detection methods that generate a rough bounding box estimate for the object and then use ICP (iterative closest points) for to refine the estimate~\cite{RamnSoria2020GraspPA} or graph cut segmentation, an all-in-one detection and segmentation approach simplifies training and can leverage GPU (Graphics Processing Unit) hardware acceleration. 

\subsubsection{From Segmentation to Point Cloud}
\begin{figure}[htbp]
          \centering
          \smallskip
          \smallskip
          \includegraphics[width=0.6\linewidth]{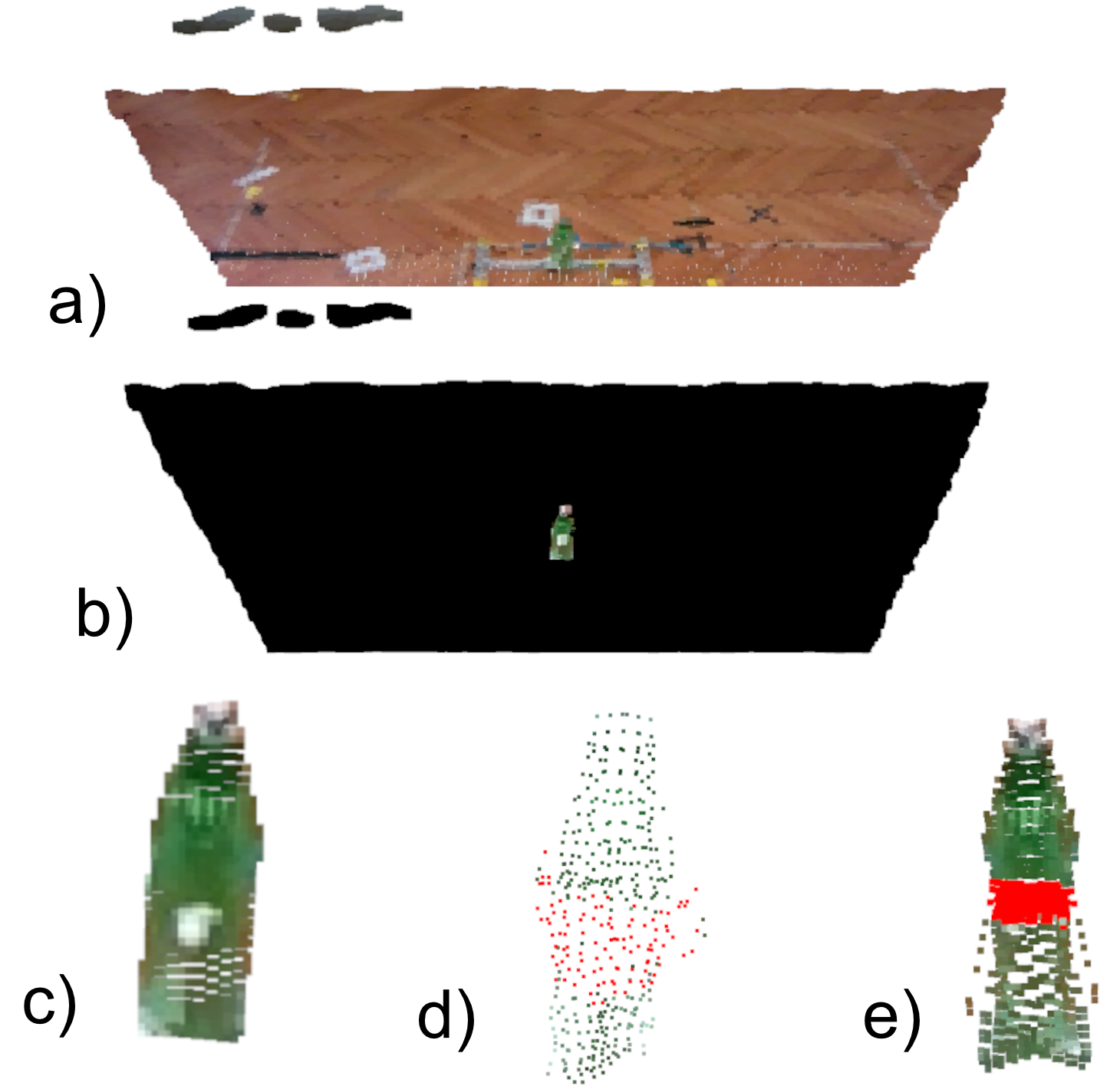}
          \\
          \centering
          \caption{The grasp planning pipeline. In a), we see the point cloud as it was created from an unmasked RGB frame and the corresponding depth frame. In b), we see a point cloud created from a RGB frame that is masked around the bottle in the center of the frame. Then, c) shows the full point cloud we get by removing all masked points and applying radius outlier removal. Finally, d) shows the downsampled point cloud fused with a copy of itself that is rotated around the main axis of the point cloud. Grasping candidates are highlighted in red, e) shows the same point cloud in full resolution for better illustration.}
          \label{grasp_planning}
\end{figure}

Once a target object is segmented from the scene, we apply the segmentation mask to the RGB image. We then combine the masked RGB image with its corresponding depth image to create a cropped point cloud of the scene. Next, we remove all points in the point cloud which are black due to having been masked out in the previous step. This preprocessing yields a point cloud of the surface of the object that is visible to the camera. Finally, we apply radius outlier removal (as implemented in \textit{Open3D}~\cite{Zhou2018}) to correct for any possible outliers that the mask might have falsely included as part of the object and apply voxel downsampling to reduce the computational load for further computations. A visual representation of this process is given in \Cref{grasp_planning}.

\subsection{Computing a Grasp}
 We choose a relatively simple geometry-based strategy for selecting grasping points which is inspired by Zapata-Impata et al.~\cite{ZapataImpata2019FastGC} who previously implemented their algorithm on a static platform using a robotic arm. Bringing this algorithm to a flying platform, we make the following assumptions that are necessary for stable flight with the quadcopter after grasping the object: the object fits into the gripper on the drone, it weighs less than \SI{500}{\gram} and is approximately axially symmetric. 
 
 Our adapted algorithm can be summarized into four major steps:
\begin{enumerate}
    \item Estimate the centroid of the point cloud and determine its pose.
    \item Duplicate the point cloud and rotate it around the axis with the largest extent by 180 degrees, making use of the assumption about symmetry to estimate front and back of the object.  
    \item Determine the new centroid and the new pose of the combined point cloud. 
    \item Determine a set of candidate points by taking the intersection of the point cloud and a cutting plane (\SI{2.5}{\centi \meter} thick) that goes through the estimated centroid and is normal to the axis with the largest extent. These candidates mark the possible contact spots for our gripper on the front and the back. 
\end{enumerate}

\section{Mobile Vision Architecture}

\begin{figure*}[htbp]
          \centering
          \smallskip
          \smallskip
          \includegraphics[width=0.85\linewidth]{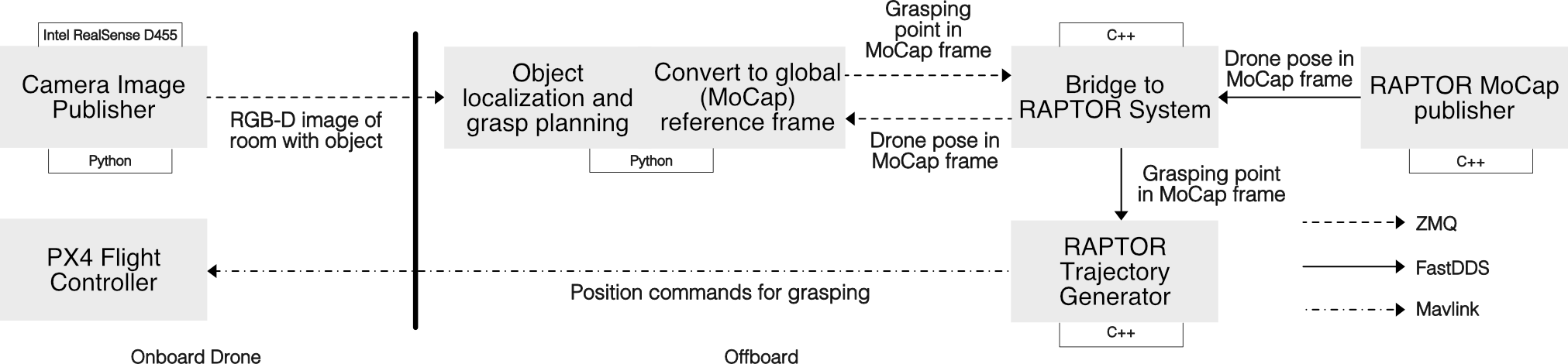}
          \\
          \centering
          \caption{The dataflow for the vision system integrated into the \textit{RAPTOR} system. Each block represents a single process. This architecture allows the vision system to be a drop-in replacement for the existing target object localization using a motion capture (MoCap) system.}
          \label{dataflow}
\end{figure*}

\subsection{Compute Platform}
The use of mobile compute platforms with powerful GPUs on drones has been explored before~\cite{Foehn2022AgiliciousOA} and has been shown to be a viable alternative to offboard computation. However, it comes at the cost of significantly increased weight and power consumption onboard the drone while still not living up to the processing times possible by performing the computations offboard. As \textit{detectron2} runs at very slow speeds on these mobile compute platforms (around 10 seconds per frame on Nvidia Jetson Nano), we choose to use run the computations on a more powerful offboard desktop PC with an \textit{Nvidia RTX 3070} GPU to accelerate segmentation runtimes. Onboard the drone, an \textit{Nvidia Jetson Nano}~\cite{NvidiaJetson} is used to capture, compress and send the RGB and depth frames from the \textit{Intel RealSense D455}~\cite{IntelRealSense} camera. 

Additionally, we have also equipped the \textit{Nvidia Jetson Nano} with an \textit{Intel AC8265} wireless networking card which is much faster than the wireless networking hardware on simpler single-board-computers like a \textit{Raspberry Pi 4}. The two computers are connected over an industrial-grade wireless router with 1~Gbps bandwidth. 

\subsection{Data Pipeline}
We send color and depth frames to the offboard computer over the local network, do the most intensive computations there and then forward the result to the \textit{RAPTOR} system over the local network as shown in \Cref{dataflow}. Not only does this architecture allow us to use the vision system as a drop-in replacement for the previously used motion capture system, but its modular nature also allows the vision system to be quickly integrated into other platforms. Additionally, the system can also be deployed on an onboard computer with ease should sufficiently powerful compute platforms be available. 

\begin{figure*}[htpb]
          \centering
          \smallskip
          \smallskip
          \includegraphics[width=0.925\linewidth]{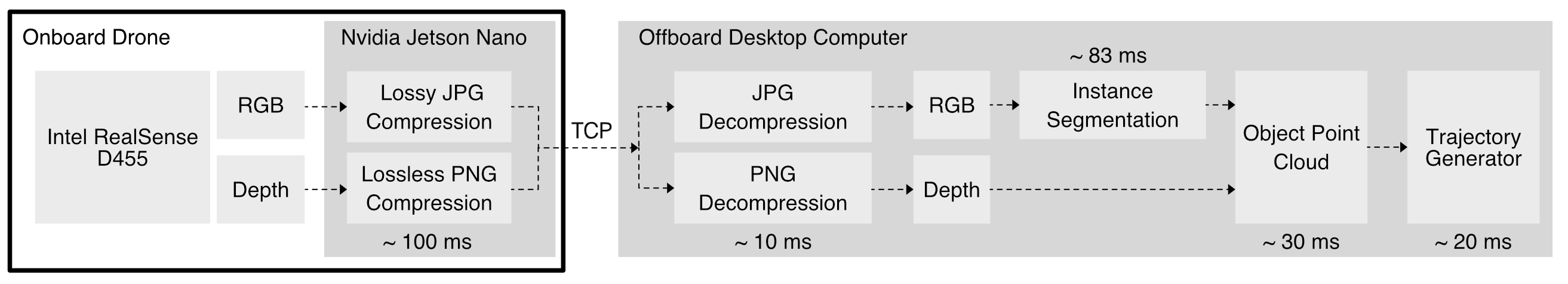}
          \\
          \centering
          \caption{Dataflow of the image streaming pipeline using hybrid compression scheme (lossy JPG for RGB images and lossless PNG for depth frames. Depth frames require lossless compression to preserve localization accuracy whereas for RGB frames, we can use computationally cheaper lossy JPG compression.}
          \label{img_streaming}
\end{figure*}

\subsubsection{Dataflow}
We start with the camera image publisher running onboard the drone on a \textit{Nvidia Jetson Nano} which compresses both RGB and depth frames from the \textit{Intel RealSense D455} camera and sends them to the offboard computer. On the offboard computer, we run scene segmentation with \textit{detectron2} \cite{wu2019detectron2} and object localization and grasp planning using \textit{Open3D}~\cite{Zhou2018}. The obtained target coordinates are transformed into the global motion capture frame, in which the trajectory planning for \textit{RAPTOR} happens. To transform the coordinates, we receive the pose of the drone from a bridge process. In turn, the bridge process receives the transformed target coordinates which it then forwards to the trajectory generator. From there, it will be sent to the flight controller of the drone (\textit{Pixhawk 4}~\cite{PX4Site}) using \textit{MAVlink}~\cite{MAVLinkSite}, which then handles all lower levels of control to fly the drone to the target position.

\subsubsection{Synchronisation of Processes}
We use a request-reply communication pattern within \textit{ZeroMQ}, ensuring synchronization of the different processes. In case one process fails, this failure will interrupt the request-reply scheme and the other processes will not keep operating on old or invalid data as it would be the case with the \textit{Fast DDS} publisher-subscriber model. 

\subsubsection{Transport and Serialization}
The \textit{RAPTOR} system is currently not using ROS (Robot Operating System)~\cite{Macenski2022RobotOS} but rather a custom \textit{Fast DDS} wrapper to reduce computational overhead by a factor of up to 1.5~\cite{Kronauer2021LatencyOO}. To enable interoperability with other languages than C++, we extended the system building up on \textit{ZeroMQ (ZMQ)}~\cite{ZMQSite} and \textit{Protobuf}~\cite{ProtobufSite}. This extends the system to be used with all other major programming languages (Python, Rust, \dots) while maintaining support for ROS-like messages and low computational overhead. As visible in \Cref{dataflow}, the vision system uses ZMQ (combined with \textit{Protobuf} messages) to connect to the \textit{RAPTOR} system while the \textit{RAPTOR} system relies on \textit{Fast DDS} and \textit{MAVLink} to enable communication in between processes.

\subsection{Image Streaming and Compression}
Real-time image streaming is one of the main challenges related to offboard computing. We cannot use lossy compression for the depth frames as that would incur large losses in the localization accuracy. For the RGB frames, however, we can use lossy compression to minimize the transmission size in a computationally efficient way. 

\subsubsection{Hybrid compression scheme} Accordingly, we make use of a hybrid compression scheme (\Cref{img_streaming}) similar to work presented by Coatsworth et al.~\cite{Coatsworth2014AHL}. We access one stream of RGB frames and one stream of Z16 depth frames from the \textit{Intel RealSense D455}~\cite{IntelRealSense} camera, both with a resolution of 640 by 480 pixels. Then, we compress the RGB frame using lossy JPG compression and the depth frame using lossless PNG compression. For JPG compression, we use a quality of 95 whereas for PNG compression, we use a compression factor of 2 (compression algorithms from \textit{OpenCV}~\cite{opencv_library} are used). 

\begin{figure}[htbp]
          \centering
          \includegraphics[width=0.65\linewidth]{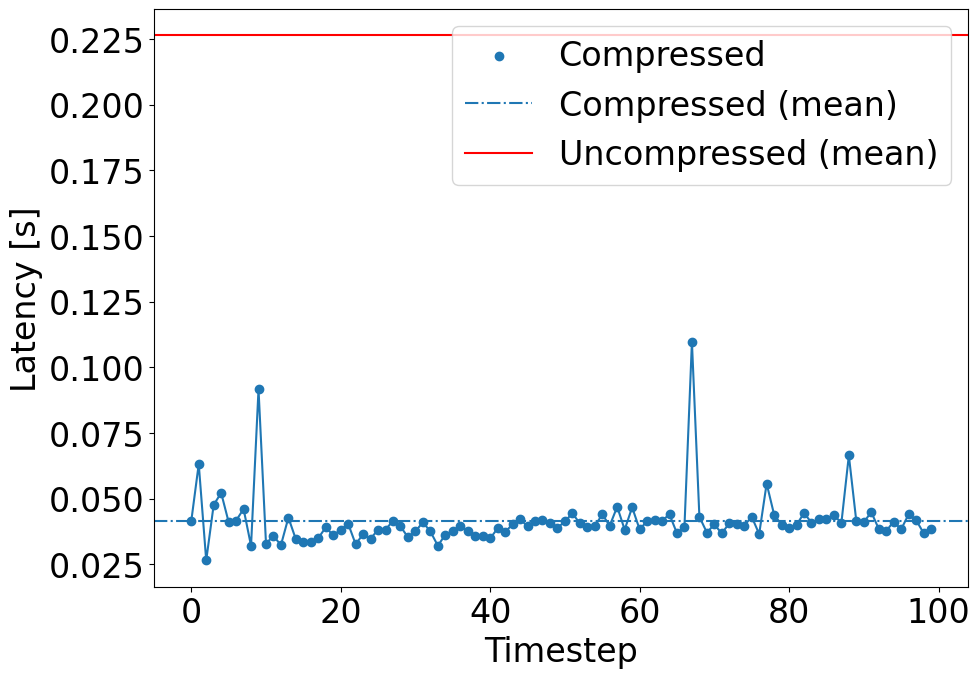}
          \\
          \centering
          \caption{The transit time for a pair of one JPG-compressed RGB frame and one PNG-compressed depth frame sent over \textit{imagezmq}, both with a resolution of 640 by 480 pixels. The mean transit time is \SI{41}{\milli\second}.}
          \label{img_latency}
\end{figure}


\subsubsection{Latency} In total, the time to compress the frames and send them to the offboard computer is much lower than directly using image segmentation on the onboard computer given the currently suitable onboard computers. Our use of offboard computing is motivated by the total time needed to process one frame; however, in more latency-critical applications or for outdoor deployment, using onboard segmentation with reduced framerates may be necessary. For possible future work exploring outdoor scenarios, we can easily port the system to a single onboard computer to take advantage of lower latencies.

\subsubsection{Compression Parameters} To find the compression parameters, we used a runtime analysis of the system. The rate at which the onboard computer sends images should at most be the rate at which the offboard computer can process them. Thus, we choose the highest compression parameters that allow the runtime to be less than that of the main loop. Runtimes are listed in \Cref{img_streaming}.

\subsection{Perfomance Analysis}

In the following, we analyze the system performance of the vision system. There are two relevant components: latency through processing and latency through network transit times. 

\subsubsection{Process Timings} \label{perf_runtimes}
The camera image publisher running on the \textit{Jetson Nano} takes \SI{100}{\milli\second} on average to get, compress, and send a pair of RGB and depth frames. Image decompression, scene segmentation and grasp planning on the offboard computer takes approximately \SI{120}{\milli\second} on average, of which the largest part can be attributed to image segmentation (\SI{83}{\milli\second}) and point cloud processing (\SI{30}{\milli\second}). The \textit{RAPTOR} processes for trajectory generation and data transfer run at a speed of \SI{20}{\milli\second} per iteration. 

These runtimes allow us to deploy the system with real-time performance, which allows us to use it in a variety of different scenarios. For swooping motions, we do not explicitly make use of the real-time capabilities due to motion blur at the presence of both high linear and angular velocities in swooping motions. However, for use cases that utilize flight maneuvers at low velocities, real-time capabilities provide constant feedback that can be used for more traditional closed-loop visual servoing.

\subsubsection{Networking Latency}
All offboard processes run on the same computer and the time messages spend in transport is on the order of nanoseconds. However, there is considerable networking latency when sending the image frames from the \textit{Jetson Nano} on the drone to the offboard computer. For transport, the TCP implementation of \textit{ZMQ} is used. A showcase of the transit times of images is given in \Cref{img_latency}.

\section{Results and Discussion}

We conducted real-world experiments with two goals: \begin{enumerate}
    \item Evaluate grasping success rates for rapid aerial swooping grasps.  
    \item Characterize the localization error of our vision-based system with respect to the motion capture system. 
\end{enumerate} 

\begin{figure}
          \centering
          \includegraphics[width=0.7\linewidth]{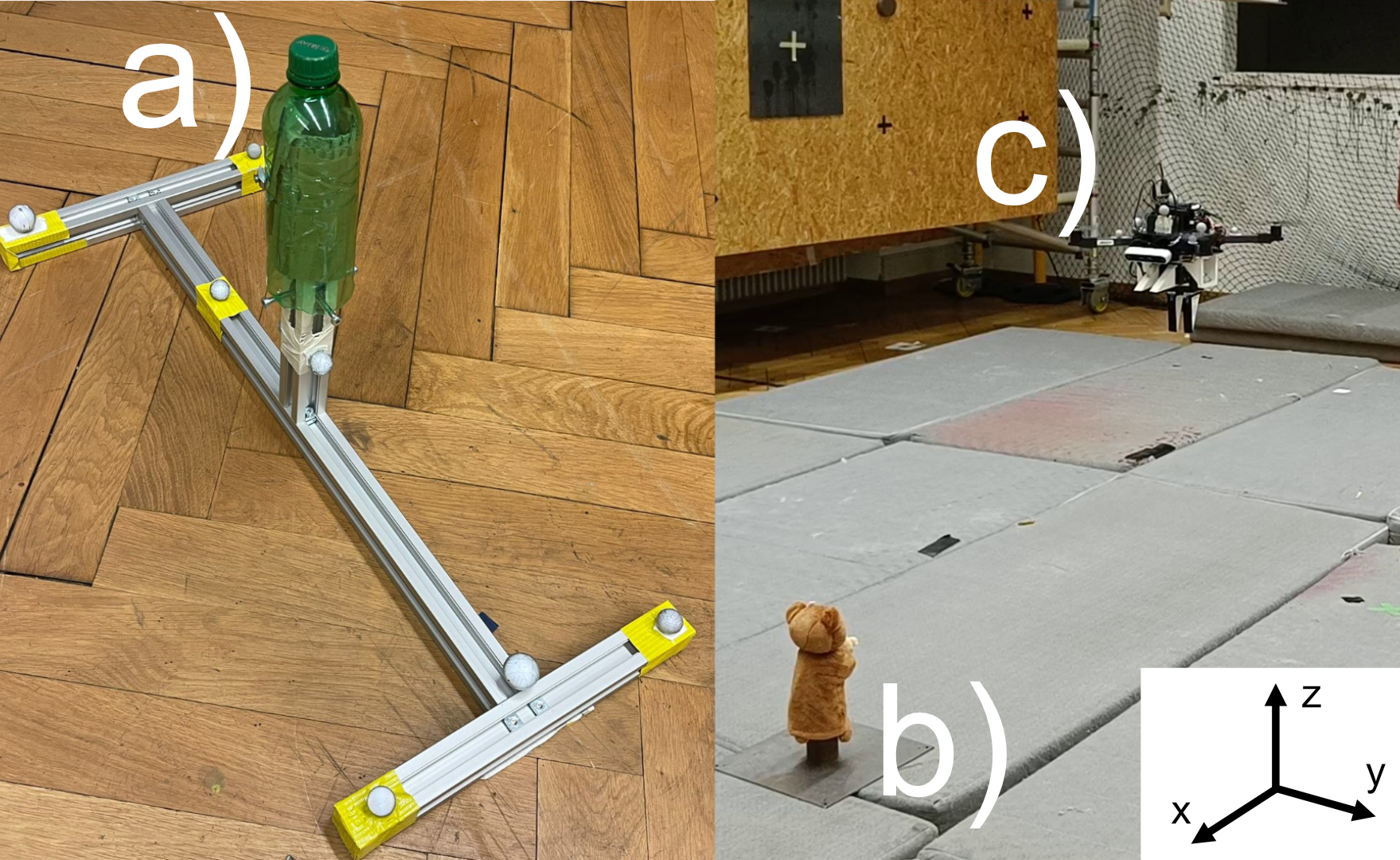}
          \\
          \centering
          \caption{The experimental setup used. Pictured: a) the bottle is mounted on a stand equipped with motion capture markers to allow us to perform localization tests b) the teddy bear is mounted without any markers for the grasping tests c) the drone is approaching the object for grasping.}
          \label{experiment_setup}
\end{figure}

\subsection{Methodology}
In the following, an overview of the methodology for both experiments is provided.

\subsubsection{Motion Capture System}

\begin{figure}[t]
          \centering
          \smallskip
          \smallskip
          \includegraphics[width=0.4\linewidth]{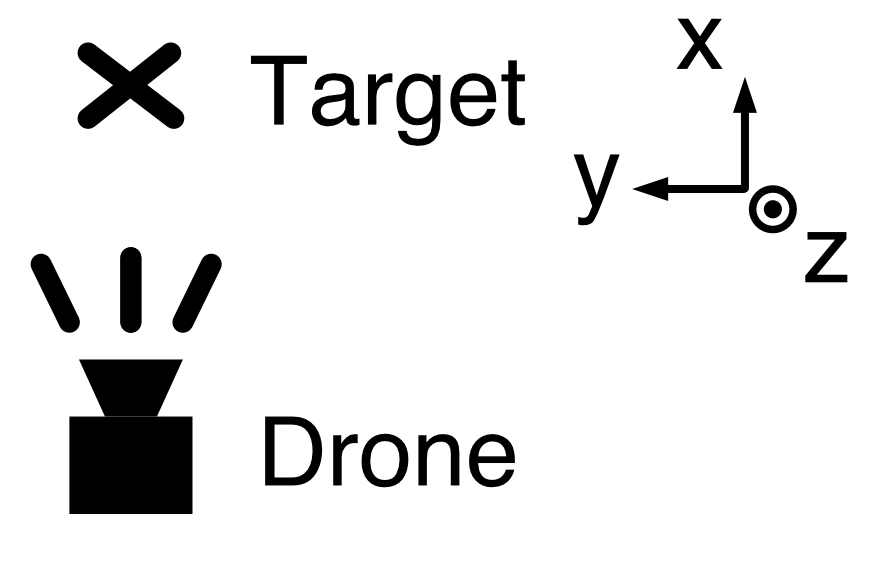}
          \\
          \centering
          \caption{Schema of the experimental setup and related coordinate system. This represents the coordinate system used in \Cref{bear_error} and \Cref{bottle_error}. In all experiments, the drone kept its rotation as shown with the camera being aligned with the x-axis.} 
          \label{experiment_schema}
\end{figure}

Both experiments we conducted took place in a motion capture room where the motion capture system was used for estimating the pose of the drone. The drone itself moves in the motion capture coordinate system, which is illustrated in \Cref{experiment_schema}. In the experiments, the camera view on the drone was always aligned with the x-axis of the motion capture system. All errors are computed in the motion capture frame.

\subsubsection{Target Objects}

\begin{figure}
          \centering
          \includegraphics[width=0.8\linewidth]{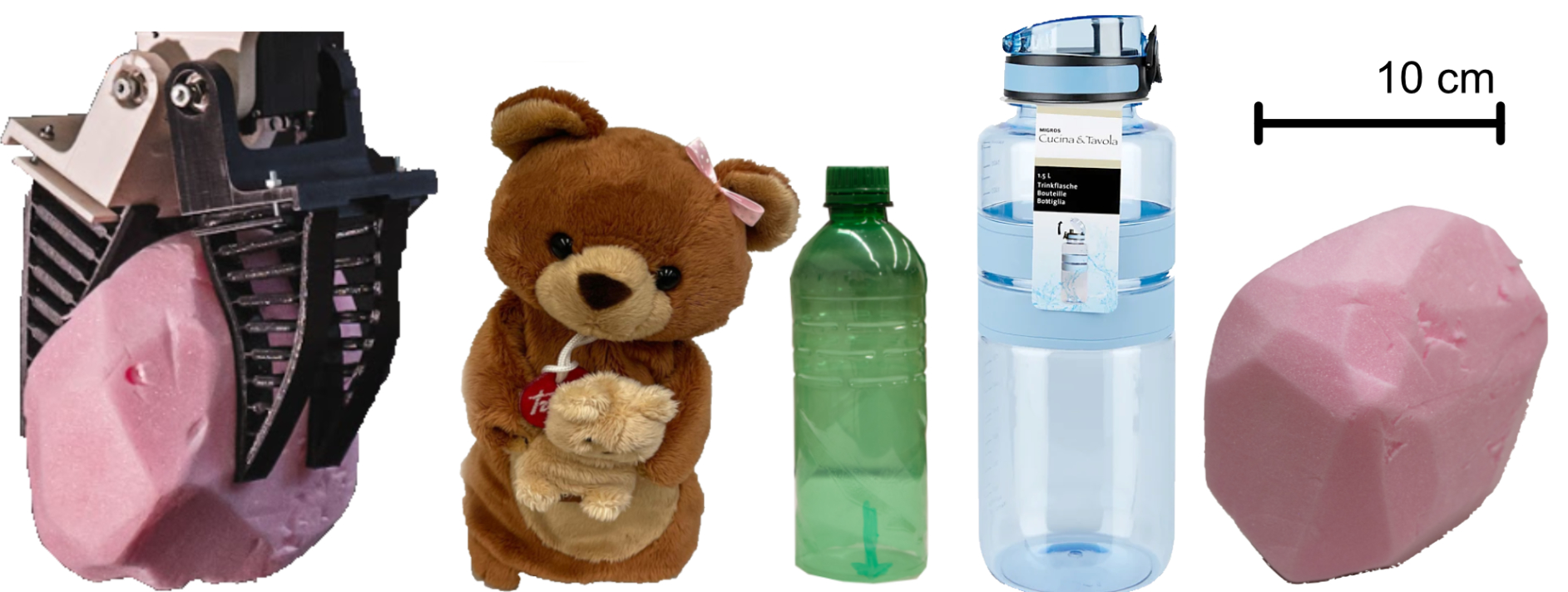}
          \\
          \centering
          \caption{From left to right: gripper grasping Styrofoam object, teddy bear, bottle for localization tests, bottle for grasping tests, Styrofoam object for baseline. Objects depicted at relative scale to each other.}
          \label{stand} 
\end{figure}

There are various relevant target objects that are displayed in \Cref{stand}. Target objects have to be in the COCO dataset that was used for training our object detector, weigh under 500 grams and fit in the gripper of the drone. 

For our experiments, we focus on the teddy bear object and the bottle object. Both of these objects fulfil all our criteria, in addition, they can stand on their own and are not blown away by the drone upon approach. The downwash generated from the drone eliminated other objects we tried such as handbags and small books. 

Furthermore, we could add distractor objects as long as they are of a different object class than the target object. The used instance segmentation framework (\textit{detectron2}) features strong performance for differentiating in between different object classes that make this technique robust to dealing with distractor objects. If the system detects multiple objects of the target object class, it would aim to grasp the first object detected. 

\subsection{Grasping Experiments}

We conducted the grasping experiments with the teddy bear test object and the larger bottle object depicted in \Cref{stand}. The test objects were placed on a metal plate approximately in the middle of the room, without any markers attached. The teddy bear object was placed on a metal weight to secure its position. For the bottle, no further measures were taken to ensure its position stayed the same. 36 grasps were attempted for each object with the following procedure: 

\begin{enumerate}
    \item Fly to the center of the room and get a view of the object.
    \item Hover for 5 seconds and gather a first grasp estimate for the target object.
    \item Reposition to execute the swooping maneuver for the grasp in a straight line. 
    \item Execute the swoop (\SI{3}{\second} total duration) and grasp the object, then drop the object and land again.
\end{enumerate}

The observed results are shown in \Cref{success_rates}. For the teddy bear test object, 89\% of grasps were successful. The bottle test object proved more challenging with a 64\% success rate. This difference in success rates can be attributed to difference in size: the bottle is smaller than the teddy bear, which places higher demands on precise grasps. However, uncertainty in trajectory tracking is substantial for dynamic aerial maneuvers as previously observed by Appius et al.~\cite{Appius2022RAPTORRA}. Therefore, grasping success rates decrease for smaller objects which the quadcopter misses the target due to imperfect trajectory tracking. In \Cref{success_rates}, we also show the percentage of measurements where the estimated y-coordinate of the object fell within the actual width of the object. We can see the percentages of samples within tolerance are close to the grasp success rates, which further suggests that object width and grasp success are correlated. 

Previous experiments using the \textit{RAPTOR} platform used a motion capture system for target localization and can be used to contextualize our results. Given the previous results in Appius et al.~\cite{Appius2022RAPTORRA}, we choose to compare our test objects to the test objects that are closest in size. The teddy bear test object is similarly sized to the styrofoam object and the bottle is comparable to the previously used bottle by Appius et al.~\cite{Appius2022RAPTORRA} (\SI{1.5}{\centi \meter} larger diameter). The previously shown success rates are 100\% for the styrofoam object and 61\% for the bottle. Taking these success rates as a baseline, the grasping performance shows no significant decrease using our marker-less system.  
 
\begin{table}
\smallskip
\smallskip
\caption{Grasping Success Rates.}
\label{success_rates}
\begin{center}
\begin{tabular}{cccc}
\toprule
Object & Vision (ours) & \thead{Estimates within\\width tolerance} & \thead{Motion Capture\\(baseline)} \\ 
\midrule
Teddy Bear &  89\% & 90\% & 100\% \\
PET Bottle &  64\% & 70\% & 61\% \\
\bottomrule
\end{tabular}
\end{center}
\end{table}

In conclusion, the achieved grasping success rates indicate the potential of our marker-less vision-based system as an alternative for the previously used marker-based localization systems. In combination with vision-based localization techniques~\cite{Forster2015SemiDirectVO, MurArtal2015ORBSLAMAV} that can be deployed on quadcopters, the drone could be used outside a motion capture space.

\subsection{Localization Accuracy Experiments}

In this set of experiments, we evaluate the localization accuracy of the system from different positions with the drone while keeping the camera view aligned with the x-axis in the coordinate system shown in \Cref{experiment_schema}. By varying the position of the drone, we characterize the distribution of the localization error with samples from a diverse set of positions. In doing so, we seek to quantitatively assess the localization performance and find possible limits and weak spots of our system. 

\begin{figure}
          \centering
          \smallskip
          \smallskip
          \includegraphics[width=0.7\linewidth]{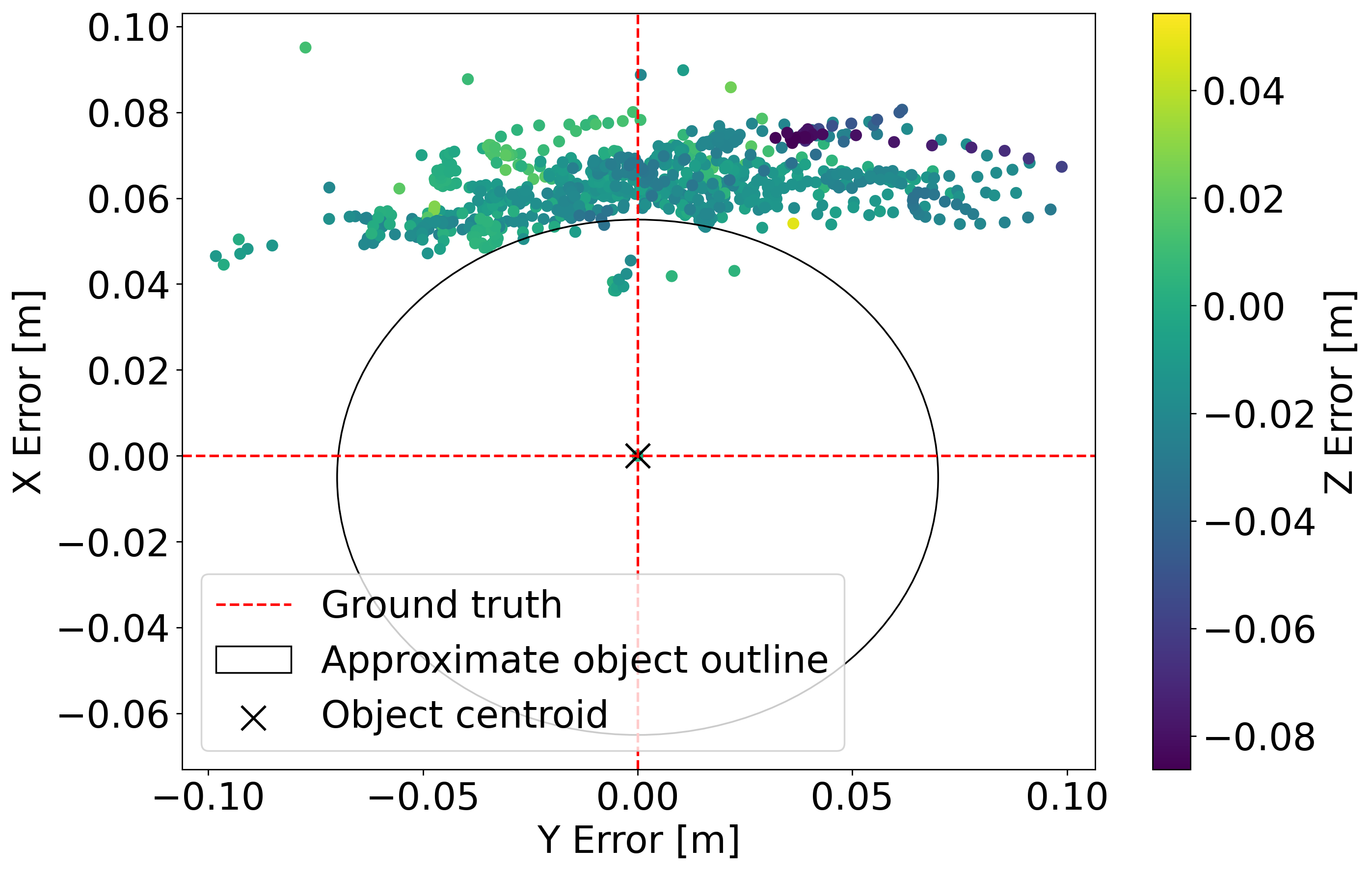}
          \\
          \centering
          \caption{Localization errors for the \textbf{teddy bear test object}. 862 measurements were taken from 42 different positions with the camera view always being aligned with the x-axis (\Cref{experiment_setup}). The slight asymmetry and thickness of the bear lead to increased errors along the x-axis, distributed around a mean squared error of \SI{6.2}{\centi \meter}. Errors in the y-axis and z-axis are distributed around mean squared errors of \SI{2.9}{\centi \meter} and \SI{1.7}{\centi\meter}.}
          \label{bear_error}
\end{figure}

\begin{figure}
          \centering
          \smallskip
          \smallskip
          \includegraphics[width=0.7\linewidth]{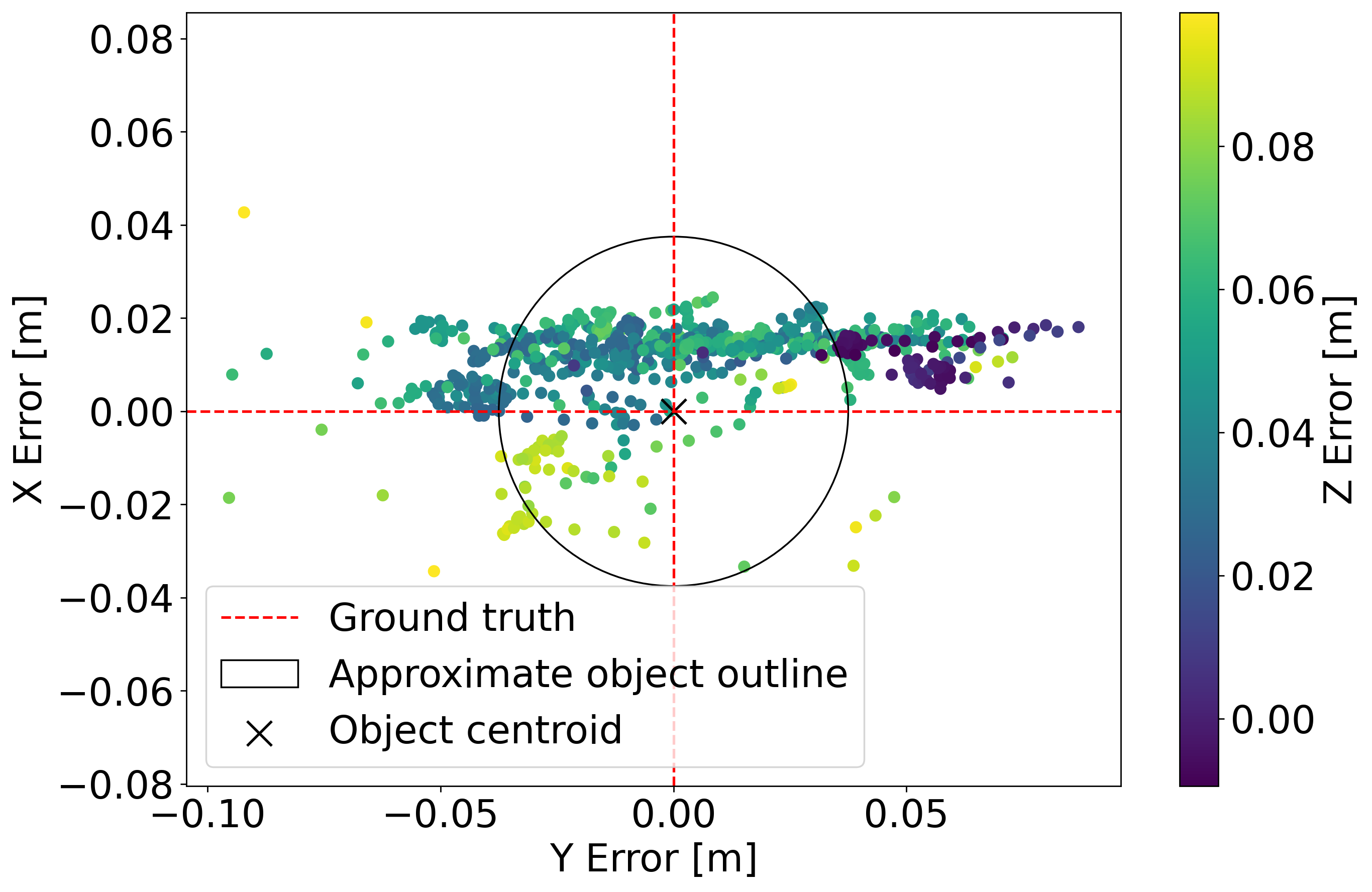}
          \\
          \centering
          \caption{Localization errors for the \textbf{bottle test object}. 766 measurements were taken from 42 different positions with the camera view always being aligned with the x-axis (\Cref{experiment_setup}). Faulty segmentation masks of the bottle lead to increased errors along the z-axis, distributed around a mean squared error of \SI{4.5}{\centi \meter}. Errors in the y-axis and z-axis are distributed around mean squared errors of \SI{1.3}{\centi \meter} and \SI{2.7}{\centi\meter}.}
          \label{bottle_error}
\end{figure}

In this experiment, the drone is moving on a grid defined by relative translation to the target object. Along the x-axis (\textbf{facing the object}) we move from \SI{0.6}{\meter} (minimum depth rating from the camera) to \SI{1.8}{\meter} away from the target object. Along the y-axis (\textbf{perpendicular to camera view}), we move from \SI{-1.5}{\meter} to \SI{1.5}{\meter} translation with respect to the target object. In total, the drone  Each detection of the target object is depicted by a point in \Cref{bear_error} and \Cref{bottle_error}.  

To retrieve the position of the object using the motion capture system, we place the target objects (teddy bear and bottle shown in \Cref{stand}) on a stand which is equipped with motion capture markers. The stand is shown in \Cref{experiment_setup}.

\subsubsection{Teddy bear test object}

For the teddy bear test object, we observed seemingly large centroid localization errors. Being slightly asymmetric, this object violates our assumption of symmetry. Furthermore, its thickness poses a challenge to our system - grasp pose estimation is based on the visible point cloud of the object. However, for thicker objects, this simplification starts to fail. To aid in estimating the centroid, we would a more complete 3D point cloud of the target object. Nonetheless, we also see that grasping success does not necessarily correlate with centroid localization accuracy, as for this test object, our system achieved a 94\% success rate when grasping. 

\subsubsection{Bottle test object}

For the bottle test object, the centroid localization errors are visibly smaller than for the teddy bear test object. Particularly, the mean squared error in the x-axis are significantly decreased (from \SI{6.2}{\centi\meter} to \SI{1.3}{\centi\meter}). This can mainly be attributed to higher symmetry and lower thickness. However, a higher error in the z-axis can be observed due to imperfect segmentation masks. Overall, while the mean squared errors are significantly lower, once again, we do not observe a correlation with the grasping success. Despite lower localization errors, only 64\% of grasps succeeded due to imperfect reference tracking of the drone near the ground. This added position uncertainty is reflected in the grasping success rates. 

Furthermore, for both objects, the lowest MSE was obtained at translations of \((x=\SI{1.2}{\meter}, y=\SI{0}{\meter})\).

\subsubsection{Tradeoff: Generalization versus Localization Precision}
The overall mean squared error (MSE) we present (i.e. the bottle test object has an MSE of \SI{5.9}{\centi\meter}) is larger compared to other methods that use \textbf{stronger prior assumptions}. For example, Ramón-Soria et al.~\cite{RamnSoria2020GraspPA} claim an MSE as low as \SI{2}{\centi \meter} using ICP to align the detected point cloud with a known 3D point cloud of the target object. This difference in precision represents a tradeoff in between generalizing to objects of many different shapes and classes (our approach) and being able to estimate the position of an object with even higher precision~\cite{RamnSoria2020GraspPA}. With our results, we particularly address the need for \textbf{higher generalization abilities without relying on priors} such as a 3D model of the target object. Especially for real-world applications such as in warehouses, higher generalization abilities will be crucial for the success of robotic manipulators. 


\section{Conclusion}
We have demonstrated our vision-based system for real-time scene segmentation and grasp planning and its capabilities as a replacement for motion-capture-based target localization solutions. Our learning-based system eliminates the need for objects to have any markers on them or be of distinctive previously known shapes. We detailed our system architecture including how we synchronized processes running at different speeds and deployed them on the \textit{RAPTOR} platform. We have shown that our system can precisely localize objects for grasping and have validated the performance in real-world grasping tests with different objects on a state-of-the-art platform for aerial grasping, showing that there is nearly no loss in grasping success compared to a motion capture baseline and demonstrating a successful transfer of geometric grasp planning to an aerial platform. This work shows how we achieved real-time vision-based target object localization and grasping and serves as a foundation for future work that will extend our system's capabilities and use case scenarios. 


For future work, pushing the boundaries of the current system will be the focus of research efforts to come. Particularly, dynamic grasping in unstructured outdoor environments using visual-inertial odometry for self-localization and the ability to grasp larger, strongly asymmetric objects are the open challenges we see as most important. For enhanced grasping abilities, generating an estimate of the full three-dimensional point cloud of the object by learning-based methods or dense mapping could improve the grasping and target localization performance and allow more extensive grasp planning. Solving these challenges will be an important step towards giving robots a true understanding of the shape of objects and how to interact with these objects in any kind of environment. Interaction based on vision is something humans and animals have mastered long ago; giving robots a similar perception of space and objects through learning will be a major achievement for any kind of autonomously acting robotic systems.

\addtolength{\textheight}{0cm}   

\bibliography{./root.bib}

\begin{thebibliography}{10}
\providecommand{\url}[1]{#1}
\csname url@samestyle\endcsname
\providecommand{\newblock}{\relax}
\providecommand{\bibinfo}[2]{#2}
\providecommand{\BIBentrySTDinterwordspacing}{\spaceskip=0pt\relax}
\providecommand{\BIBentryALTinterwordstretchfactor}{4}
\providecommand{\BIBentryALTinterwordspacing}{\spaceskip=\fontdimen2\font plus
\BIBentryALTinterwordstretchfactor\fontdimen3\font minus
  \fontdimen4\font\relax}
\providecommand{\BIBforeignlanguage}[2]{{%
\expandafter\ifx\csname l@#1\endcsname\relax
\typeout{** WARNING: IEEEtran.bst: No hyphenation pattern has been}%
\typeout{** loaded for the language `#1'. Using the pattern for}%
\typeout{** the default language instead.}%
\else
\language=\csname l@#1\endcsname
\fi
#2}}
\providecommand{\BIBdecl}{\relax}
\BIBdecl

\bibitem{GarridoJurado2014AutomaticGA}
S.~Garrido-Jurado, R.~Mu{\~n}oz-Salinas, F.~J. Madrid-Cuevas, and M.~J.
  Mar{\'i}n-Jim{\'e}nez, ``Automatic generation and detection of highly
  reliable fiducial markers under occlusion,'' \emph{Pattern Recognit.},
  vol.~47, pp. 2280--2292, 2014.

\bibitem{IntelRealSense}
\BIBentryALTinterwordspacing
``Intel realsense d455,'' accessed: 2022-09-15. [Online]. Available:
  \url{https://web.archive.org/*/https://www.intelrealsense.com/depth-camera-
  d455}
\BIBentrySTDinterwordspacing

\bibitem{8237584}
K.~He, G.~Gkioxari, P.~Dollár, and R.~Girshick, ``Mask r-cnn,'' in \emph{2017
  IEEE International Conference on Computer Vision (ICCV)}, 2017, pp.
  2980--2988.

\bibitem{wu2019detectron2}
Y.~Wu, A.~Kirillov, F.~Massa, W.-Y. Lo, and R.~Girshick, ``Detectron2,''
  \url{https://github.com/facebookresearch/detectron2}, 2019.

\bibitem{Zhou2018}
Q.-Y. Zhou, J.~Park, and V.~Koltun, ``{Open3D}: {A} modern library for {3D}
  data processing,'' \emph{arXiv:1801.09847}, 2018.

\bibitem{Appius2022RAPTORRA}
A.~X. Appius, E.~Bauer, M.~Blöchlinger, A.~Kalra, R.~Oberson, A.~Raayatsanati,
  P.~Strauch, S.~Suresh, M.~von Salis, and R.~K. Katzschmann, ``Raptor: Rapid
  aerial pickup and transport of objects by robots,'' in \emph{2022 IEEE/RSJ
  International Conference on Intelligent Robots and Systems (IROS)}, 2022, pp.
  349--355.

\bibitem{Olson2011AprilTagAR}
E.~Olson, ``Apriltag: A robust and flexible visual fiducial system,''
  \emph{2011 IEEE International Conference on Robotics and Automation}, pp.
  3400--3407, 2011.

\bibitem{Lippiello2016HybridVS}
V.~Lippiello, J.~Cacace, A.~Santamaria-Navarro, J.~Andrade-Cetto, M.~A.
  Trujillo, Y.~Rodr{\'i}guez, and A.~Viguria, ``Hybrid visual servoing with
  hierarchical task composition for aerial manipulation,'' \emph{IEEE Robotics
  and Automation Letters}, vol.~1, pp. 259--266, 2016.

\bibitem{Buonocore2015HybridVS}
L.~R. Buonocore, J.~Cacace, and V.~Lippiello, ``Hybrid visual servoing for
  aerial grasping with hierarchical task-priority control,'' \emph{2015 23rd
  Mediterranean Conference on Control and Automation (MED)}, pp. 617--623,
  2015.

\bibitem{Fishman2021ControlAT}
J.~Fishman and L.~Carlone, ``Control and trajectory optimization for soft
  aerial manipulation,'' \emph{2021 IEEE Aerospace Conference (50100)}, pp.
  1--17, 2021.

\bibitem{Fishman2021DynamicGW}
J.~Fishman, S.~Ubellacker, N.~Hughes, and L.~Carlone, ``Dynamic grasping with a
  "soft" drone: From theory to practice,'' \emph{2021 IEEE/RSJ International
  Conference on Intelligent Robots and Systems (IROS)}, pp. 4214--4221, 2021.

\bibitem{Luo2020NaturalFV}
B.~Luo, H.~Chen, F.~Quan, S.~Zhang, and Y.~hui Liu, ``Natural feature-based
  visual servoing for grasping target with an aerial manipulator,''
  \emph{Journal of Bionic Engineering}, vol.~17, pp. 215--228, 2020.

\bibitem{Lin2019AutonomousVA}
L.~Lin, Y.~Yang, H.~Cheng, and X.~Chen, ``Autonomous vision-based aerial
  grasping for rotorcraft unmanned aerial vehicles,'' \emph{Sensors (Basel,
  Switzerland)}, vol.~19, 2019.

\bibitem{RamnSoria2020GraspPA}
P.~Ram{\'o}n-Soria, B.~C. Arrue, and A.~Ollero, ``Grasp planning and visual
  servoing for an outdoors aerial dual manipulator,'' \emph{Engineering},
  vol.~6, pp. 77--88, 2020.

\bibitem{Besl1992AMF}
P.~J. Besl and N.~D. McKay, ``A method for registration of 3-d shapes,''
  \emph{IEEE Trans. Pattern Anal. Mach. Intell.}, vol.~14, pp. 239--256, 1992.

\bibitem{Soria2017DetectionLA}
P.~R. Soria, B.~C. Arrue, and A.~Ollero, ``Detection, location and grasping
  objects using a stereo sensor on uav in outdoor environments,'' \emph{Sensors
  (Basel, Switzerland)}, vol.~17, 2017.

\bibitem{LoweDavid2004DistinctiveIF}
G.~LoweDavid, ``Distinctive image features from scale-invariant keypoints,''
  \emph{International Journal of Computer Vision}, 2004.

\bibitem{Rosten2008FasterAB}
E.~Rosten, R.~B. Porter, and T.~Drummond, ``Faster and better: A machine
  learning approach to corner detection,'' \emph{IEEE Transactions on Pattern
  Analysis and Machine Intelligence}, vol.~32, pp. 105--119, 2008.

\bibitem{Thomas2016VisualSO}
J.~R. Thomas, G.~Loianno, K.~Daniilidis, and V.~R. Kumar, ``Visual servoing of
  quadrotors for perching by hanging from cylindrical objects,'' \emph{IEEE
  Robotics and Automation Letters}, vol.~1, pp. 57--64, 2016.

\bibitem{Seo2017AerialGO}
H.~Seo, S.~Kim, and H.~J. Kim, ``Aerial grasping of cylindrical object using
  visual servoing based on stochastic model predictive control,'' \emph{2017
  IEEE International Conference on Robotics and Automation (ICRA)}, pp.
  6362--6368, 2017.

\bibitem{ZapataImpata2019FastGC}
B.~S. Zapata-Impata, P.~Gil, J.~Pomares, and F.~Torres, ``Fast geometry-based
  computation of grasping points on three-dimensional point clouds,''
  \emph{International Journal of Advanced Robotic Systems}, vol.~16, 2019.

\bibitem{Haschke2021GeometryBasedGP}
R.~Haschke, G.~Walck, and H.~J. Ritter, ``Geometry-based grasping pipeline for
  bi-modal pick and place,'' \emph{2021 IEEE/RSJ International Conference on
  Intelligent Robots and Systems (IROS)}, pp. 4002--4008, 2021.

\bibitem{Jiang2011EfficientGF}
Y.~Jiang, S.~Moseson, and A.~Saxena, ``Efficient grasping from rgbd images:
  Learning using a new rectangle representation,'' \emph{2011 IEEE
  International Conference on Robotics and Automation}, pp. 3304--3311, 2011.

\bibitem{Mousavian20196DOFGV}
A.~Mousavian, C.~Eppner, and D.~Fox, ``6-dof graspnet: Variational grasp
  generation for object manipulation,'' \emph{2019 IEEE/CVF International
  Conference on Computer Vision (ICCV)}, pp. 2901--2910, 2019.

\bibitem{Levine2016EndtoEndTO}
S.~Levine, C.~Finn, T.~Darrell, and P.~Abbeel, ``End-to-end training of deep
  visuomotor policies,'' \emph{ArXiv}, vol. abs/1504.00702, 2016.

\bibitem{He2017MaskR}
K.~He, G.~Gkioxari, P.~Doll{\'a}r, and R.~B. Girshick, ``Mask r-cnn,''
  \emph{IEEE Transactions on Pattern Analysis and Machine Intelligence},
  vol.~42, pp. 386--397, 2017.

\bibitem{Bolya2019YOLACTRI}
D.~Bolya, C.~Zhou, F.~Xiao, and Y.~J. Lee, ``Yolact: Real-time instance
  segmentation,'' \emph{2019 IEEE/CVF International Conference on Computer
  Vision (ICCV)}, pp. 9156--9165, 2019.

\bibitem{Wang2022YOLOv7TB}
C.-Y. Wang, A.~Bochkovskiy, and H.-Y.~M. Liao, ``Yolov7: Trainable
  bag-of-freebies sets new state-of-the-art for real-time object detectors,''
  \emph{ArXiv}, vol. abs/2207.02696, 2022.

\bibitem{Jocher_YOLOv5_by_Ultralytics_2020}
\BIBentryALTinterwordspacing
G.~Jocher, ``{YOLOv5 by Ultralytics},'' 5 2020. [Online]. Available:
  \url{https://github.com/ultralytics/yolov5}
\BIBentrySTDinterwordspacing

\bibitem{Foehn2022AgiliciousOA}
P.~Foehn, E.~Kaufmann, A.~Romero, R.~Pěni{\v{c}}ka, S.~Sun, L.~Bauersfeld,
  T.~M. Laengle, G.~Cioffi, Y.~Song, A.~Loquercio, and D.~Scaramuzza,
  ``Agilicious: Open-source and open-hardware agile quadrotor for vision-based
  flight,'' \emph{Science Robotics}, vol.~7, 2022.

\bibitem{NvidiaJetson}
\BIBentryALTinterwordspacing
``Nvidia jetson nano,'' accessed: 2022-09-15. [Online]. Available:
  \url{https://web.archive.org/*/https://www.nvidia.com/en-us/autonomous-
  machines/embedded-systems/jetson-nano/}
\BIBentrySTDinterwordspacing

\bibitem{PX4Site}
\BIBentryALTinterwordspacing
``Pixhawk,'' accessed: 2022-09-15. [Online]. Available:
  \url{https://web.archive.org/*/https://pixhawk.org/}
\BIBentrySTDinterwordspacing

\bibitem{MAVLinkSite}
\BIBentryALTinterwordspacing
``Mavlink,'' accessed: 2022-09-15. [Online]. Available:
  \url{https://web.archive.org/*/https://mavlink.io/}
\BIBentrySTDinterwordspacing

\bibitem{Macenski2022RobotOS}
S.~Macenski, T.~Foote, B.~P. Gerkey, C.~Lalancette, and W.~Woodall, ``Robot
  operating system 2: Design, architecture, and uses in the wild,''
  \emph{Science Robotics}, vol.~7, 2022.

\bibitem{Kronauer2021LatencyOO}
T.~Kronauer, J.~Pohlmann, M.~Matth{\'e}, T.~Smejkal, and G.~P. Fettweis,
  ``Latency overhead of ros2 for modular time-critical systems,'' \emph{ArXiv},
  vol. abs/2101.02074, 2021.

\bibitem{ZMQSite}
\BIBentryALTinterwordspacing
``Zeromq,'' accessed: 2022-09-15. [Online]. Available:
  \url{https://github.com/zeromq/}
\BIBentrySTDinterwordspacing

\bibitem{ProtobufSite}
\BIBentryALTinterwordspacing
``Protobuf,'' accessed: 2022-09-15. [Online]. Available:
  \url{https://github.com/protocolbuffers/protobuf}
\BIBentrySTDinterwordspacing

\bibitem{Coatsworth2014AHL}
M.~Coatsworth, J.~Tran, and A.~Ferworn, ``A hybrid lossless and lossy
  compression scheme for streaming rgb-d data in real time,'' \emph{2014 IEEE
  International Symposium on Safety, Security, and Rescue Robotics (2014)}, pp.
  1--6, 2014.

\bibitem{opencv_library}
G.~Bradski, ``{The OpenCV Library},'' \emph{Dr. Dobb's Journal of Software
  Tools}, 2000.

\bibitem{Forster2015SemiDirectVO}
C.~Forster, Z.~Zhang, M.~Gassner, M.~Werlberger, and D.~Scaramuzza, ``Svo:
  Semidirect visual odometry for monocular and multicamera systems,''
  \emph{IEEE Transactions on Robotics}, vol.~33, no.~2, pp. 249--265, 2017.

\bibitem{MurArtal2015ORBSLAMAV}
R.~Mur-Artal, J.~M.~M. Montiel, and J.~D. Tard{\'o}s, ``Orb-slam: A versatile
  and accurate monocular slam system,'' \emph{IEEE Transactions on Robotics},
  vol.~31, pp. 1147--1163, 2015.

\end{thebibliography}
\bibliographystyle{IEEEtran}

\end{document}